\documentclass[conference, a4paper]{IEEEtran}

\IEEEoverridecommandlockouts
\IEEEpubid{\makebox[\columnwidth]{\textbf{979-8-3315-6655-5/25/\$31.00 ©2025 IEEE}\hfill}
\hspace{\columnsep}\makebox[\columnwidth]{}}
\usepackage[english]{babel}
\usepackage[utf8]{inputenc}
\usepackage[T1]{fontenc}
\usepackage{cite}
\usepackage{xcolor} 

\usepackage[pdftex]{graphicx}
\usepackage[cmex10]{amsmath}
\usepackage{multirow}
\usepackage{array}
\usepackage{svg}
\usepackage[lofdepth,lotdepth]{subfig}
\usepackage{graphicx}
\usepackage{amssymb}
\usepackage{caption}
\usepackage{tikz}
\usepackage{hyperref} 
\usepackage{algorithm}
\usepackage{algpseudocode}
\usetikzlibrary{automata, positioning, arrows}
\usepackage{times}
\usepackage{titlesec}

\begin{document}

\title{Guided Decoding and Its Critical Role in Retrieval-Augmented Generation}

\author{
    \IEEEauthorblockN{
        Özgür Uğur, Musa Yılmaz, Esra Şavirdi,\\
        Özay Ezerceli, Mahmut El Huseyni, Selva Taş, Reyhan Bayraktar
    }
    \IEEEauthorblockA{
        Newmind AI\\
        Istanbul, Türkiye\\
        \texttt{\{ougur, myilmaz, esavirdi, oezerceli, mehussieni, stas, rbayraktar\}@newmind.ai}
    }
}

\maketitle

\begin{abstract}
The integration of Large Language Models (LLMs) into various applications has driven the need for structured and reliable responses. A key challenge in Retrieval-Augmented Generation (RAG) systems is ensuring that outputs align with expected formats while minimizing hallucinations. This study examines the role of guided decoding in RAG systems, comparing three methods, Outlines, XGrammar, and LM Format Enforcer, across different multi-turn prompting setups (0-turn, 1-turn, and 2-turn). By evaluating success rates, hallucination rates, and output quality, we provide insights into their performance and applicability. Our findings reveal how multi-turn interactions influence guided decoding, uncovering unexpected performance variations that can inform method selection for specific use cases. This work advances the understanding of structured output generation in RAG systems, offering both theoretical insights and practical guidance for LLM deployment.
\end{abstract}


\section{Introduction}
The rapid rise of Large Language Models (LLMs) has transformed natural language processing, enabling applications across diverse domains such as question-answering, content generation, and conversational systems. However, a persistent challenge lies in ensuring that LLM outputs adhere to specific structural formats, a critical requirement for practical applications like data integration, API compatibility, and automated workflows. Retrieval-Augmented Generation (RAG), introduced by \cite{lewis2021rag}, enhances LLMs by incorporating external knowledge retrieval, thereby improving factual accuracy and contextual relevance. Although RAG addresses some limitations of standalone LLMs, it does not inherently guarantee structured output, which remain essential to meet user-defined constraints in real-world scenarios. Recent research underscores this gap; authors in \cite{Liu_2024} highlight the industry’s growing demand for user-centered, restricted LLM outputs.

To bridge this gap, guided decoding backends have emerged as a promising solution, restricting LLM output to predefined formats or grammars. These methods leverage techniques such as finite-state machines, pushdown automata, or character-level enforcement to ensure compliance with structural requirements. For example, authors introduces Outlines \cite{willard2023efficientguidedgenerationlarge}, a method that employs finite-state machines for efficient and structured text generation. Similarly, other approaches such as XGrammar \cite{dong2024xgrammarflexibleefficientstructured} and LM Format Enforcer \cite{Noamgat} offer flexible mechanisms to enforce complex formats like JSON or domain-specific schemas, enhancing LLM utility in structured contexts.
\section{Related Work}
The development of RAG and guided decoding methods builds on a rich foundation of research aimed at enhancing the capabilities of LLMs. This section reviews prior studies relevant to our investigation, focusing on RAG, the challenge of structured outputs, and the guided decoding techniques evaluated in this study.

\textbf{Structured Outputs in LLMs.} The importance of structured outputs has gained increasing attention as LLMs are deployed in practical settings requiring specific formats, such as JSON, YAML, or domain-specific schemas. Industry insights \cite{Liu_2024} highlight that unconstrained outputs often fall short in tasks like data processing and API integration. Therefore, enforcing structural constraints is essential, making guided decoding a critical area of study.

\textbf{Guided Decoding Methods.} Guided decoding backends constrain LLM outputs to predefined formats, addressing the limitations of unconstrained generation. One notable approach is Outlines, proposed in \cite{willard2023efficientguidedgenerationlarge}, which leverages finite-state machines (FSMs) to guide text generation efficiently, ensuring compliance with regular grammars while maintaining computational scalability. This method is particularly suited for applications that require predictable structured output.

For complex structures like context-free grammars (e.g., JSON or code), XGrammar uses pushdown automata to enforce syntax, offering flexibility and precision. In contrast, LM Format Enforcer \cite{Noamgat} applies strict character-level constraints, with its practical use detailed in Gat's publicly available implementation despite lacking a formal paper. While guided decoding enforces strict structure, it can be computationally intensive. To address this limitation, speculative methods such as Ranked Speculative Decoding (RSD) improve efficiency by using draft models and reward-based token selection, speeding up generation without sacrificing quality, especially for long texts \cite{liao}.

\textbf{Monitor-Guided Decoding for Code Completion.}  
Modular Guided Decoding (MGD) uses static analysis to improve code generation, boosting compilation rates and enabling smaller models to outperform larger ones. It generalizes across languages and coding constraints \cite{Agrawal}.

\section{Guided Decoding Impact on RAG Performance}
\subsection{Experiment Setup}

We conducted experiments using a high-performance inference engine powered by vLLM, with \textbf{xgrammar} as the default backend for guided decoding. This setup is designed for structured output generation, such as JSON schema or regex-based decoding. Other supported backends include \textbf{lm format enforcer} and \textbf{outlines}.

Initially, vLLM v0 was employed for its compatibility with structured decoding frameworks. vLLM v1 introduced issues, restricting decoding to the \texttt{xgrammar:no\_fallback} mode, which generates errors with unsupported schemas. Consequently, vLLM v0 remains our preferred implementation. Future updates to vLLM v1 are expected to address these limitations.

The experiment utilized the OpenAI-compatible server models Qwen2.5-72B-Instruct and LLaMA-3.3-70B-Instruct. The setup involved four stages:

\begin{enumerate}
    \item \textbf{Retrieving Relevant Documents}: Using RAG to extract query-specific context.
    \item \textbf{Defining Model Input}: Setting structured instructions, prompts, and response formats.
    \item \textbf{Configuring Guided Decoding}: Testing Outlines, XGrammar, and LM Format Enforcer under identical conditions.
    \item \textbf{Evaluating Multi-Turn Conversations}: Assessing performance across 0-turn, 1-turn, and 2-turn scenarios.
\end{enumerate}

\subsection{\textbf{MultiTurn Algorithm }}
\begin{algorithm}
\caption{MultiTurn RAG Eval}
\label{alg:multiturn-rag-eval}
\begin{algorithmic}[1]
\Function{MultiTurnEval}{$dataset$, $n$}
\State $chat\_hist \gets [system prompt]$
\For{$j \gets 1$ to $n$} \Comment{add n example turns}
    \State $usr\_ex \gets \text{"rag ctx: \{ctx\} {}query: \{q\}"}$
    \State $asst\_ex \gets \text{"resp: \{r\} {}doc ids: \{truth\_id\}"}$
    \State $chat\_hist \gets chat\_hist + [usr\_ex, asst\_ex]$
\EndFor

\State $usr\_eval \gets \text{"rag ctx: \{ctx\} {}query: \{q\}"}$ 
\State $chat\_hist \gets chat\_hist + [usr\_eval]$
\State $model\_resp \gets \text{GetModelResp}(chat\_hist)$
\State $resp\_ids \gets \text{ExtractIDs}(model\_resp)$ \Comment{regex}
\State $result \gets \text{Eval}(truth\_id, resp\_ids)$

\State \textbf{return} $result$
\EndFunction

\Function{Eval}{$truth\_ids$, $resp\_ids$}
\State $corr \gets [i \text{ for } i \text{ in } resp\_ids \text{ if } i \text{ in } truth\_ids]$
\State $fp \gets [i \text{ for } i \text{ in } resp\_ids \text{ if } i \text{ not in } truth\_ids]$
\State $tp \gets (|corr| > 0 \text{ and } |fp| = 0)$
\State \textbf{return} $\{$
\State \hspace{\algorithmicindent} $"success":$ $tp,$
\State \hspace{\algorithmicindent} $"hallucination":$ $|fp| > 0$
\State $\}$
\EndFunction
\end{algorithmic}
\end{algorithm}

The algorithm \ref{alg:multiturn-rag-eval} describes the multi-turn RAG evaluation process, illustrating how it operates at different levels of depth of conversation.

The algorithm takes as input a dataset of contexts, queries, and reference document identifiers, with parameter n specifying the number of exemplar turns. For \textbf{n=0}, it uses only the system prompt and the evaluation query; for \textbf{n=1,2}, it prepares the corresponding exemplar exchanges demonstrating the expected discourse patterns and the reference citation methodology.

The framework constructs a conversation history with exemplar exchanges followed by the evaluation query. For each query, it retrieves the corresponding RAG context containing ground truth document identifiers. The model response is evaluated by comparing its extracted document references with the identifiers in the RAG context. A response is considered successful when it references at least one correct document identifier while avoiding hallucinated references.

This framework enables systematic analysis of how conversation history depth affects retrieval accuracy and hallucination rates, providing insights into multiturn RAG system behavior.

\subsection{\textbf{Guided Decoding Methods}}

\subsubsection{\textbf{FSM-Based Outlines}} 
This approach leverages finite-state machines (FSMs) for efficient text generation,  guaranteeing structural validity with \(O(1)\) complexity per token. It is especially well-suited to domains that require strict syntactic or semantic constraints, such as legal and technical documentation.

The comments on the algorithm suggest its application in a broader context, particularly in parsing. By applying \hyperref[alg:fsm-sub-sequences]{Algorithm 1} to each string in a set \( V \) using combined FSMs for each parse state, it becomes possible to determine parser configurations. These configurations include the Pushdown Automaton (PDA) states, corresponding FSM states, and potential terminal symbols. The analogy extends to using the pre-image of the PDA’s transition map to identify PDA stack values that can read the PDA states \( q \in Q \) and terminal symbol sets \( V \) of a parser configuration.

\textbf{FSM Representation of Constraints.}
Outlines represent regular expressions and context-free grammars (CFGs) as finite-state machines, where states correspond to valid prefixes of a structured sequence. The FSM tracks valid transitions, determining which tokens can legally follow a given sequence. This eliminates the need for exhaustive vocabulary filtering. 

\textbf{Efficient Vocabulary Indexing.}
To accelerate constraint enforcement, Outlines precomputes a mapping from FSM states to valid tokens. This mapping denoted as $\sigma: Q \to \mathcal{P}(V)$, enables constant-time token validity checks. Unlike naive approaches that iterate over all vocabulary tokens per step, Outlines retrieves valid tokens in $O(1)$ time on average. \cite{lew2023sequential}

\textbf{Token Sampling with FSM Constraints.}
During inference, the Outlines method modifies token sampling by applying FSM constraints to ensure structured outputs. The FSM tracks the current state and dynamically determines the valid token set. The next token is sampled from a constrained probability distribution, adhering to structural rules. This method is applicable to various formats, such as floating-point numbers, programming syntax, and structured data like JSON and XML.

\subsubsection{\textbf{XGrammar (Pushdown Automata-Based)}}
XGrammar is a high-performance engine that accelerates LLMs by $100\times$ using precomputed token masks, a persistent execution stack, and parallel grammar processing, supporting real-time generation with broad compatibility.

\textbf{Vocabulary Partitioning and Token Mask Optimization:}  
Tokens are classified as context-independent or context-dependent, with an adaptive cache to reduce memory and speed up validation.

\textbf{Persistent Execution Stack:}  
Manages parsing states efficiently, enabling fast branching and minimal memory overhead.

\textbf{Pushdown Automata Optimization:}  
Improves CFG parsing by inlining rules and reducing ambiguity.

\textbf{Parallel Mask Generation with LLM Inference:}  
Runs grammar processing parallel to GPU-based inference, minimizing latency.

\subsubsection{\textbf{LM Format Enforcer}}
LM Format Enforcer ensures adherence to predefined formats by filtering token probabilities, allowing only compliant tokens. It integrates with local LMs to improve reliability and consistency. Unlike rigid methods, it offers flexible enforcement that preserves the model's formatting style, dynamically evaluating valid token sequences to balance compliance with autonomy, ensuring high output quality.

\subsection{\protect\textbf{Dataset}}
The dataset used in this study contains metadata spanning multiple dialogue turns. Although we report a total of 750 samples, only 507 are publicly accessible on Hugging Face owing to privacy restrictions. The dataset can be accessed via our Hugging Face repository \footnote{\url{https://huggingface.co/datasets/newmindai/siu-rag-data}}.

\begin{table}[h]
\caption{Dataset Overview Across Different Turns}
\centering
\footnotesize
\begin{tabular}{|l|c|c|c|}
\hline
\textbf{Metric} & \textbf{0-Turn} & \textbf{1-Turn} & \textbf{2-Turns} \\
\hline
Total Ref. & 4909 & 2482 & 1622 \\
Unique Ref. & 3614 & 1955 & 1310 \\
Total Samples & 750 & 375 & 250 \\
\hline
\end{tabular}
\label{tab:dataset_overview}
\end{table}

\section{Results and Discussion}
We present result graphs that illustrate the differences between the three guided decoding methods.

Our implementation addressed several challenges specific to \textbf{Turkish legal documents.} The specialized \texttt{(doc\_id)document\_id(/doc\_id)} format required precise enforcement, with all methods showing significant improvement in handling these complex structures as conversation turns increased. Despite the complexity of agglutinative Turkish morphology and specialized legal vocabulary, all guided decoding approaches maintained high semantic quality (judge scores consistently higher than 91).  Complex references in Turkish legal documents (e.g., \texttt{"344.0321.DOR.2021\_1630505603\_page\_623"})  were increasingly well-handled in multi-turn scenarios, with false positive rates dropping dramatically from hundreds to single digits.

\begin{table}[h]
\caption{E2E Generation Time per Sample (sec)}
\centering
\begin{tabular}{|l|c|c|}
\hline
\textbf{Backend} & \textbf{LLaMA-3.3-70B-Instruct} & \textbf{Qwen2.5-72B-Instruct} \\
\hline
Outlines           & \textbf{30.642} & 50.766 \\
\hline
XGrammar           & \textbf{30.282} & 50.784 \\
\hline
LM Format Enforcer & \textbf{30.534} & 51.468 \\
\hline
\end{tabular}
\label{tab:backend_comparison}
\end{table}
As shown in Table \ref{tab:backend_comparison}, LLaMA-3.3-70B-Instruct processes fewer tokens per sample than Qwen2.5-72B-Instruct, which supports larger inputs and produces more extensive outputs in multiturn contexts. This indicates that Qwen2.5-72B-Instruct is optimized for longer, more complex queries, whereas LLaMA-3.3-70B-Instruct delivers faster responses on simpler tasks. Furthermore, as multiturn complexity grows, generation time increases proportionally. In these scenarios, LLaMA-3.3-70B-Instruct remains more time-efficient per sample, while Qwen2.5-72B-Instruct maintains higher throughput with larger token sets.

\begin{figure}[h]
    \centering
    \includegraphics[width=0.48\textwidth, height=8.3cm, keepaspectratio]{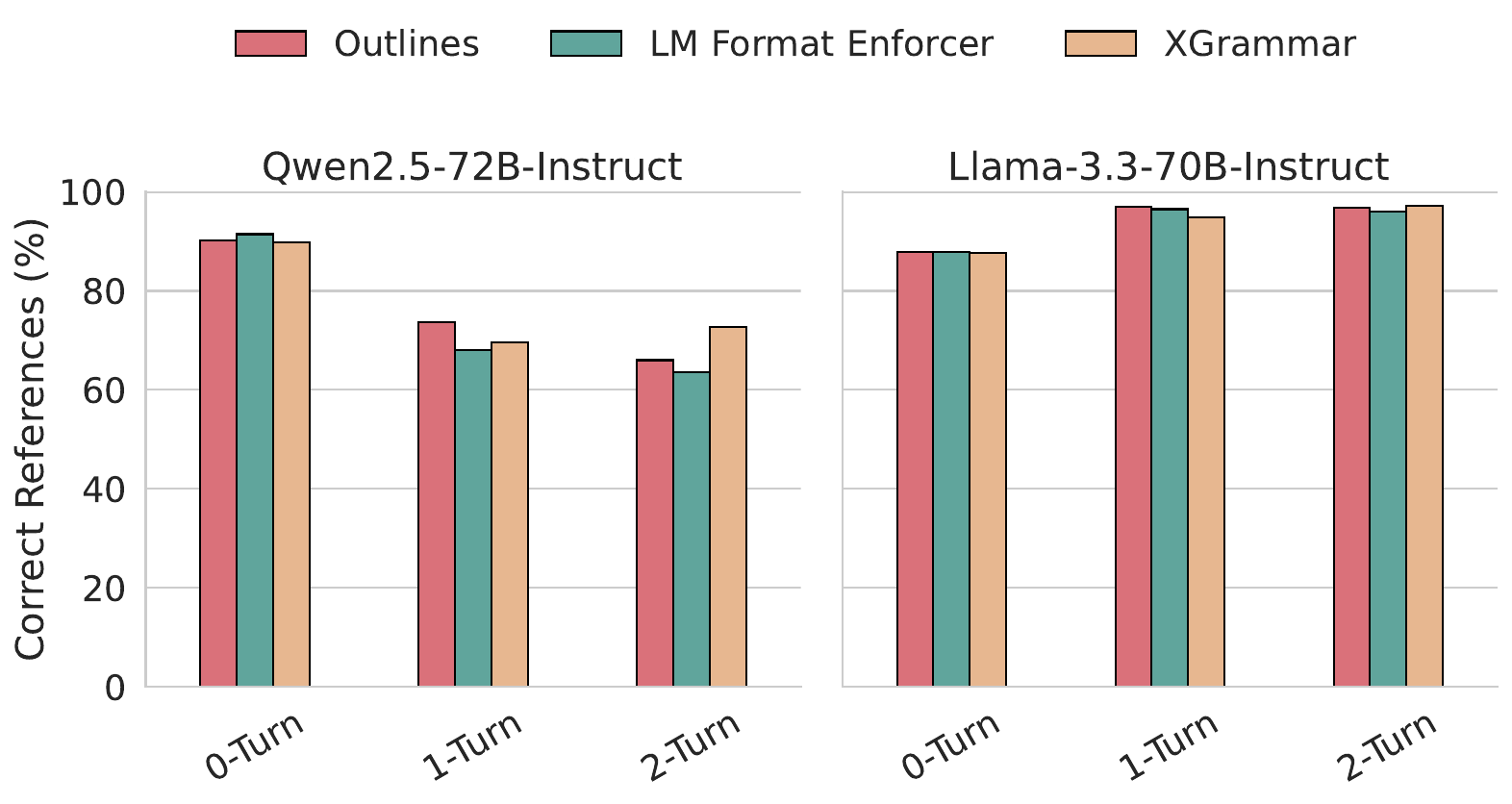}
    \caption{Performance of guided decoding backends across multi-turn scenarios}
    \label{fig:zero_turn}
\end{figure}

\textbf{Structured Output Performance Across Conversational Scenarios.} 
Table~\ref{tab:false_positive_rates} and Figure~\ref{fig:zero_turn} present a comparative analysis of false positive rates for guided decoding methods across 0-, 1-, and 2-turn conversational settings. In zero-turn interactions, LM Format Enforcer (LMF) consistently achieved the lowest false positive rates (\textbf{0.49\%} for Qwen2.5-72B-Instruct, \textbf{3.06\%} for Llama-3.3-70B-Instruct), outperforming Outlines and XGrammar. As conversational depth increased, all methods demonstrated improved performance, with LMF maintaining superior robustness: achieving the lowest rates in 1-turn (\textbf{0.73\%} and \textbf{0.33\%}) and 2-turn (\textbf{0.30\%} and \textbf{0.06\%}) contexts for Qwen2.5-72B-Instruct and Llama-3.3-70B-Instruct, respectively. Notably, Outlines and XGrammar exhibited marked gains in multi-turn settings, particularly with Qwen2.5-72B-Instruct, underscoring the benefits of conversational context in reducing structural output errors.

\begin{table}[h]
\caption{False Positive Rates of Guided Decoding}
\centering
\footnotesize
\begin{tabular}{|l|l|c|c|c|}
\hline
\textbf{Model} & \textbf{Turns} & \textbf{Outlines} & \textbf{XGrammar} & \textbf{LMF} \\
\hline
\multirow{3}{*}{Qwen2.5-72B-Instruct} 
    & 0-Turn & 0.65\% & 0.61\% & \textbf{0.49\%} \\
    & 1-Turn  & \textbf{0.32\%} & 0.41\% & 0.73\% \\
    & 2-Turns & 0.18\% & \textbf{0.12\%} & 0.30\% \\
\hline
\multirow{3}{*}{Llama-3.3-70B-Instruct} 
    & 0-Turn & 3.20\% & 3.08\% & \textbf{3.06\%} \\
    &  1-Turn  & \textbf{0.24\%} & 0.53\% & 0.33\% \\
    & 2-Turns & 0.48\% & 0.31\% & \textbf{0.06\%} \\
\hline
\end{tabular}
\label{tab:false_positive_rates}
\end{table}

Few-turn prompting significantly improved reliability, particularly in 1-turn scenarios, where explicit examples clarified the desired output structure. Outlines and XGrammar benefited the most, while LM Format Enforcer struggled with the added complexity of 2-turn prompting. Among the methods, Outlines balanced flexibility and enforcement effectively, XGrammar offered strong performance and efficiency, and LM Format Enforcer ensured strict structural compliance but often at the cost of usability. Together, guided decoding and few-turn prompting complement each other, ensuring structured and factual outputs in RAG systems.

In a RAG scenario with \textbf{10 million} chunks and \textbf{100,000 unique queries}, each potentially linked to \textbf{distinct references}, the decoding algorithm significantly affects reference accuracy.

For \textbf{Qwen2.5-72B-Instruct}, replacing \textbf{LM Format Enforcer} with \textbf{XGrammar} resulted in \textbf{1,600} additional missed references in the \textit{zero-turn} setting and \textbf{4,000} more in the \textit{one-turn} case. The performance drop suggests that XGrammar's format handling introduces notable degradation in single-turn and multi-turn contexts.

Similarly, for \textbf{LLaMA-3.3-70B-Instruct}, XGrammar led to \textbf{134} additional misses in zero-turn and \textbf{2,000} in one-turn, relative to the top-performing decoding strategies such as Outlines and LM Format Enforcer.
These findings show that in large-scale RAG systems, decoding strategy is critical for ensuring factual consistency and reference accuracy. Default decoding methods can fall short in high-recall tasks, leading to grounding errors. Outlines propose enhanced control over the output, which can substantially improve the fidelity of references.

\section{Limitations} 

\textbf{Regex \& Character Support}: Outlines limited regex support, lacking advanced features, restricts complex text processing. Its character constraints also hinder non-ASCII handling, reducing applicability in multilingual domains.

\textbf{Generation Flexibility}: Unlike LM Format Enforcer Outlines does not accommodate beam search or batched generation, reducing flexibility for tasks that require varied output‐sampling strategies \cite{vLLM-Blog}. Additionally, its partial JSON potentially yielding suboptimal outputs when slight deviations are acceptable \cite{Li}.

\textbf{Structured-Output Adaptability.}: The absence of optional-field support in JSON outputs further limits adaptability in production workflows that require flexible structured-generation methods.

Moreover, effective grammar enforcement should follow the model’s reasoning to ensure logical outputs. While tools like XGrammar and Outlines offer immediate syntax checks, LM Format Enforcer lacks support for delayed enforcement aligned with generation dynamics. XGrammar’s reliance on manual rule specification and its computational overhead underline the need for more adaptive, lightweight grammatical‐enforcement approaches.

\section{Conclusion}
Guided decoding is critical for reliable LLM deployments. This study underscores the importance of combining structured prompting with guided decoding to optimize RAG systems. Integrating external retrieval with decoding strategies that ensure format adherence enhances factual accuracy and structural reliability. Multi-turn prompting further improves control over generation while maintaining consistency with application-specific requirements. These findings highlight the need for structured decoding and prompting to advance LLM accuracy, usability, and integration in high-stakes applications.

This study explored the impact of decoding strategies on reference accuracy in large language models, comparing XGrammar with flexible methods like Outlines and LM Format Enforcer. Experiments on Qwen2.5-72B-Instruct and LLaMA-3.3-70B-Instruct show that adaptive strategies significantly reduce \textbf{reference loss}, especially in multi-turn settings.

\section*{Acknowledgment}
This study is supported by \textbf{GSI Attorney Partnership}. The authors would also like to express their gratitude for the valuable insights and support provided throughout the research process.

\renewcommand{\refname}{References}

\end{document}